\pdfoutput=1

\documentclass[11pt]{article}

\usepackage{acl}

\usepackage{times}
\usepackage{latexsym}

\usepackage[T1]{fontenc}

\usepackage[utf8]{inputenc}

\usepackage{microtype}

\usepackage{amsmath}
\usepackage{graphicx}
\usepackage{multicol}
\usepackage{adjustbox}
\usepackage{makecell}
\usepackage{tikz,graphics,float,epsf,caption,subcaption}
\usepackage{multirow}
\usepackage{dirtytalk}
\usepackage{listings}
\usepackage{fancyhdr}
\definecolor{commentsColor}{rgb}{0.497464, 0.75, 0.497464}
\usepackage{xcolor}
\usepackage{dblfloatfix}

\usepackage{tcolorbox}
\newenvironment{chat}{
    \begin{tcolorbox}[
        colback=white, 
        colframe=black, 
        coltext=black, 
        arc=5pt, 
        title={Coherence: 2 $|$ Engagingness: 4 $|$ Humanness: 3}, 
        fonttitle=\bfseries\small, 
        sharp corners, 
        left=10pt,right=10pt,top=5pt,bottom=5pt, 
    ]
    
}{
    \end{tcolorbox}
}
\usepackage[framemethod=TikZ]{mdframed}
\mdfdefinestyle{MDSMSRIGHT}{%
    linecolor=gray,
    fontcolor=black,
    outerlinewidth=0pt,
    roundcorner=5pt,
    innerrightmargin=5pt,
    innerleftmargin=5pt,
    font=\small,
    backgroundcolor=gray!50!white}
    
\mdfdefinestyle{MDSMSLEFT}{%
    linecolor=gray,
    fontcolor=black, 
    outerlinewidth=0pt,
    roundcorner=5pt,
    font=\small,
    innerrightmargin=5pt,
    innerleftmargin=5pt,
    backgroundcolor=white}

\newcommand{\user}[1]{
\hspace*{10pt}
    \begin{minipage}{0.95\linewidth}
        \begin{mdframed}[style=MDSMSRIGHT] #1 \end{mdframed}
        \vspace*{0.5pt}
    \end{minipage}\par
}

\newcommand{\model}[1]{
    \begin{minipage}{0.95\linewidth}
        \begin{mdframed}[style=MDSMSLEFT] #1 \end{mdframed}
        \vspace*{0.5pt}
    \end{minipage}\par
}
\newcommand{\translate}[1]{\small\textcolor{blue}{\textit{(#1)}}}

\title{Role-Play Zero-Shot Prompting with Large Language Models for Open-Domain Human-Machine Conversation}

\author{Ahmed Njifenjou \and Virgile Sucal \and Bassam Jabaian \and Fabrice Lefèvre  \\
        Laboratoire Inforamitque d'Avignon (LIA), Avignon Université \\
        \texttt{\{ahmed-ndouop.njifenjou \& firstname.lastname\}@univ-avignon.fr} \\}

\fancypagestyle{firstpage}{
    \fancyhf{}
    \fancyhead[L]{Latest version of a paper originally submitted to SIGDIAL \textbf{2023}}
}

\begin{document}
\thispagestyle{firstpage}
\maketitle
\begin{abstract}
 Recently, various methods have been proposed to create open-domain conversational agents with Large Language Models (LLMs). These models are able to answer user queries, but in a one-way Q\&A format rather than a true conversation. Fine-tuning on particular datasets is the usual way to modify their style to increase conversational ability, but this is expensive and usually only available in a few languages. In this study, we explore role-play zero-shot prompting as an efficient and cost-effective solution for open-domain conversation, using \textbf{capable} multilingual LLMs~\cite{beeching2023stackllama} trained to obey instructions. We design a prompting system that, when combined with an instruction-following model - here Vicuna~\cite{vicuna2023} - produces conversational agents that match and even surpass fine-tuned models in human evaluation in French in two different tasks.
\end{abstract}

\section{Introduction}

Since their introduction, Transformers~\cite{vaswani2017attention} and derivative Large Languages Models (LLMs) have constantly improved the state-of-the-art on several Natural Language Processing (NLP) tasks. Among them, is open-domain dialogue which consists in a conversational agent designed to interact socially with users on any topic while displaying human abilities, like empathy, personality or entertainment~\cite{walker2021}. Common approaches using LLMs mostly relied on fine-tuning with specific datasets often targeting one or more of these skills (e.g. PersonaChat~\cite{Zhang2018d}, Blended Skill Talk~\cite{smith-etal-2020-put}, Empathetic Dialogues~\cite{rashkin2019towards} \textit{inter alia}). These datasets are expensive to build and often available only in one language. In addition recent developments in the field of NLP have focused on LLMs trained to follow instructions~\cite{ouyang2022training}. They have the native ability to respond to users' inputs in a natural language manner. However, the leader-to-follower single-sided relationship is strongly present and they hardly display human conversational abilities straightaway.

In this work, as an attempt to deal with these issues while escaping from fine-tuning and its data dependency, we propose to use role-play via zero-shot prompting to leverage instruction-following models' abilities. This approach pertains to the newly devised prompt-based learning (PBL) paradigm~\cite{pbl_review21}. We assess this approach on two tasks, a general Persona task based on the PersonaChat dataset to bring these models to have a persona while displaying human skills and a particular case, the INT task~\cite{Riou2019b} where the speakers have to discuss an image, simulating a situated multi-modal conversation.

\section{Related Work}
\label{sec:sota}
Open-domain dialogue has seen a lot of developments with most solutions focusing mainly on fine-tuning with specific collected data. Among others we have the BlenderBot's ~\cite{roller-etal-2021-recipes} series and other closed-sourced models as Meena~\cite{adiwardana2020humanlike}, LaMDA~\cite{thoppilan2022lamda}, etc. These models display great conversational skills, but in addition to data dependency, they are often only available in English.
Foundational models are a recent trend in the field of NLP as they display multilingual abilities and state-of-the-art performance on several benchmarks. One of them, LLaMA~\cite{touvron2023llama, touvron2023llama2} is the backbone LLM of this study. These series of models are trained on an amount of data surpassing the scaling law of~\cite{hoffmann2022training}. Such capable LLMs are a prerequisite to build instruction following models.
From LLaMA resulted, among others: Alpaca~\cite{alpaca}, StackLLaMA\cite{beeching2023stackllama}, Guanaco~\cite{dettmers2023qlora} and the main model of this study, Vicuna~\cite{vicuna2023}. The latter has been fine-tuned on the ShareGPT\footnote{\url{https://sharegpt.com}} corpus which is basically a dataset of conversations produced by ChatGPT~\cite{ChatGPT} users. They are collected using a web browser plugin installed by users willing to participate in the dataset collection.
The resulting models have open-domain \textbf{responding} abilities -- which we dissociate from \textbf{conversational} abilities -- and implicitly OpenAI original model's restrictions, resulting from the ChatGPT alignment process, as described in the OpenAI documentation~\cite{Leike2023}.

The idea behind instruction following models lays within a new paradigm in NLP, coined as Prompt-Based Learning (PBL)~\cite{pbl_review21}. While using such models to perform prediction tasks, their inputs are  modified following a global template into a textual string prompt that has some unfilled slots, and then the language model is used to fill the unfilled information to obtain a final string, from which the final output can be derived, in a generative way. One major asset of the method, which makes it very powerful, is that it allows the LLM to be pre-trained on huge quantity of text and, by defining a new prompt scheme, the model can perform few-shot --- or even zero-shot --- learning, and adapt to new scenarii with few or no labelled data.

\cite{santu2023teler} proposed a general prompt taxonomy to unify complex tasks bench-marking which shares the idea of prompt structuring with our work. However, with role-play prompting, we specifically target dialogue which requires even more complex abilities. 
The role-play prompt scheme proposed in this GitHub repository~\footnote{\url{https://github.com/teknium1/alpaca-roleplay-discordbot}} intended for an LLM-based Discord bot is closer to our needs. 
However, it is designed for providing only persona-based information.
Role-play prompting as we want to demonstrate is not just limited to character or persona level instructions. Role-Play can be used in order to enforce other conversational skills, such as empathy or engagingness, which help balance the dialogue between the user and the bot.
We applied this approach to two scenarii, without fine-tuning.

\section{Methodology}
\label{sec:methodology}

\subsection{Instruction-Following vs Dialogue Skills}

Quoting~\cite{reitz2019}, \say{dialogue is more than just having a conversation. Genuine dialogue describes a way of interacting that is mutual, relational, attentive, and meaningful}. Instruction-following models, even those optimized for conversation fall short from fulfilling some of these aspects of a genuine dialogue. 
Indeed, social aspects - mutuality, relationality and attention --
especially are poorly displayed. Talking of mutuality, these systems often converse in a leader-follower structure where they are the follower and the user the leader -- this is for instance evidenced by the words used to designate user entries in ChatGPT release blog: \say{queries}, \say{instructions}~\cite{ChatGPT}.
Regarding relationality and attention, these systems lack of straightforward engagingness and personality consistency.

Commonly, these limitations have been tackled with finetuning. However, this is costly and data-dependant which data are scarce in languages other than English. As results, we formalize a general role-play prompt structure which is a more efficient and less expensive approach. Indeed, it avoids finetuning and rely on \textbf{multilingual LLMs }prompted in \textbf{English}, with external \textbf{English data} if needed (for instance persona from PersonaChat Dataset) and performs dialogue task in a \textbf{desired target language} here in \textbf{French}.

\begin{figure*}[!b]
	\centering
 \resizebox{\textwidth}{!}{
        \includegraphics[scale=1]{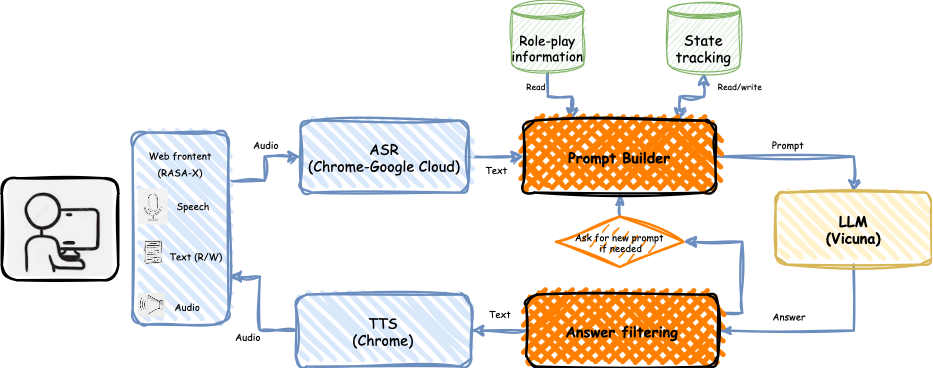}
    }
 	\caption[Architecture]{Architecture of the Role-play Zero-shot prompted dialogue system.}
	\label{fig:Architecture}
\end{figure*}
\subsection{Role-Play Prompting}
\label{subsec:roleplay-promptengineering}
It is important to understand that Role-Play Prompting here is not restricted to playing a given character. To better understand that, let's consider the simulacra and simulator framing in~\cite{shanahan2023roleplay}. LLM is a simulator which swallowed myriads of simulacra during the pre-training and at each simulation it kind of randomly selects among any of them which one\textbf{S} to display. The simulacrum is, as a matter of fact, not only about persona background~\cite{smith-etal-2020-put} but also \textit{thinking} and \textit{writing} styles, personal situation~\cite{smith-etal-2020-put}, target language (for a multilingual simulator) and information processing (long-term memory, user personalization, response filtering etc.). All possible simulacra already exist in the simulator (LLM) but it doesn't display all of them natively, in fact it can't. Role-Prompting enters the chat to make the LLM favor simulacra that are suitable for a given dialogue task. In this paper, we derived two distinct dialogue tasks to assess the efficiency of this approach.
The first task, referred as Persona task, uses Role-Play Prompting for enhancing humanness in conversation skills.
The second one, referred as INT task, derives Role-Play to allow the LLM to talk about a simulacrum instead of interpreting it.

\subsection{Prompt Structure}

Open-domain dialogue belonging to the realm of complex tasks~\cite{santu2023teler} makes the endeavour of role-play prompting more challenging. As a matter of fact, small variations in a prompt may hamper the model's observed performance. For this reason, it is mandatory to define a general prompt structure that can be adapted later to different conversational tasks. Hence, with the concerns of dealing with the previously mentioned limitations we retain the following sections -- each focusing on different aspects useful to a dialogue -- as building blocks of a prompt engineering module:

\begin{enumerate}

   \item \textbf{System Instructions} $\mathcal{I}_{s} =\{i_{s, k}\}_{k=1}^{N_{i}}$: where $N_i$ is the number of instructions $i_{s,k}$ which sharply define the target task's specifications and the global desired behaviour. This may include \textit{thinking} (inferring) and \textit{writing} (generation)  styles.

    \item \textbf{Situational context} $\mathcal{C}^{t} =\{c_{k}\}_{k=1}^{N_{c}}$: each $c_k$ is a context information that may help the model better perform the desired task. As such, it evolves with time depending on the conversation flow. It can include personality information, image and scene description, summaries of old turns or information from external sub-modules.
    
    \item \textbf{Response Instructions} $\mathcal{I}_{a} =\{i_{a,k}\}_{k=1}^{N_{i}}$: These are final instructions to incite the LLM to respond to the users' utterance with emphasis on the writing style, the target responding language and creativity with section 2) in mind.
    
    \item \textbf{Conversation History} $\mathcal{X}^{t}$: this part contains previous messages from the user ($x^{t}$) and the LLM ($y^{t}$). They can be truncated to the $k$ latest conversation turns to fit in the LLM token size limit or help the LLM focus on latest part of the conversation. In this case a summary of the $k$ removed turns $\tilde{\mathcal{X}^{k}}$ generated by an external module (also using PBL with an LLM) can be added in section 2). Conversation history, therefore becomes: $\mathcal{X}^{t-1} = \{ (x^{t-k}, y^{t-k}), ..., (x^{t-1}, y^{t-1})\} $

\end{enumerate}

These sections can be further precised into subsections and their order is set to vary as it may be suitable to give more or less importance to one section than another for the final model's response depending on the task  at hand. This will be showcased in our two experimented tasks later on. Finally, the prompt builder returns:
$
\label{eq:prompt_structure}
{
    \ensuremath{
    \mathcal{P}_{task}^t= 
    \sigma_{task} (\mathcal{I}_{s},\mathcal{C}^t, \mathcal{I}_{a}, \mathcal{X}^{t-1})
    }
}
$ where $\sigma_{task}$ is the most suitable permutation for the dialogue task at hand.
Therefore at each turn, the model maximizes the following probability: $ p(y^t|x^t,  \mathcal{P}_{task}^t)$ to respond.

\section{Experiments}
\label{sec:experiments}

All experiments are carried out in \textbf{French} 
but the prompt contains instructions mainly in \textbf{English}, one of which specifies the target response language. Given the results obtained by~\cite{lai-etal-2023-okapi} and those we present in Appendix~\ref{tab:llm-Benchamark}, we assume that this can be applied to other languages in which the assessed model performs comparably, as in French. 

\subsection{System Architecture}
\label{subsec:architecture}

The system shown in Figure~\ref{fig:Architecture} is a pipeline of several modules. These include a web interface based on the Rasa~X~\cite{rasax} tool, modified to integrate voice functionality using the Google Chrome Speech-to-text and Text-to-speech APIs. This enables users to exchange easily with the agent, either by voice (recommended) or by text. Next, a module that constructs the prompt according to the general structure described in~\ref{subsec:roleplay-promptengineering} from the user's (textual) message and information both external and internal to the conversation. After generation, if the LLM responses are not valid, a filtering module is used to apply corrections before sending them to the user (cf. Annex~\ref{sec:filtering-appendix}).

\subsection{Open-domain Conversation With Human Capabilities: the PersonaChat Task}
\label{subsec:persona-experiments}

This task involves enhancing LLM's conversational capabilities by using roles built from personality traits drawn from the \textbf{PersonaChat dataset}~\cite{zhang-etal-2018-personalizing} embedded in $\mathcal{C}^{t}$ as external information (see conversation example in Appendix~\ref{conversation:open-domain-example}). 

\subsubsection{Shallow Prompt}

Given in Appendix~\ref{subsec:persona-prompt}, it is close to Vicuna's basic prompt (Appendix~\ref{subsec:vicuna-prompt}). However, there are some additions for the sake of fair comparisons: system instructions to describe the task, contextual information (notably personality traits), and the instruction to complete the conversation history.

\subsubsection{Advanced Prompt} 

It exactly follows the structure in Section~\ref{subsec:roleplay-promptengineering} i.e. $\sigma_{task} = \mathbf{I}\mathbf{d}_{4}$ (see appendix~\ref{subsec:persona-prompt}).  
The context $\mathcal{C}^{t}$ includes the \textbf{specification of humanity} where personality traits are added with the injunction to choose a name consistent with them if necessary. 
External \textbf{modules} can augment this under certain conditions, as in~\cite{Shuster2022BlenderBot3A}. Indeed, to prevent prompts from exceeding the LLM's maximum context size while retaining the content of the entire conversation, old exchanges are summarized in a few sentences. To personalize the user experience, a line containing user-specific information is added and updated regularly. This aspect is generally referred to as long-term memory. These "modules" are actually prompts sent to an LLM. The \textbf{history $\mathcal{X}^{t}$ of the conversation} is kept at the end of the Advanced Prompt so that the system, when generating a response, has an overview of the entire conversation. 

\subsection{Simulating Multi-modal Conversations: the INT Task}
\label{subsec:int-experiments}

If a model is capable of interpreting a role in order to embody a character, we can assume that it is also capable of talking about a role without interpreting it. To test the validity of this hypothesis, we propose a prompt designed to enable the LLM to converse about a specific topic. This topic is defined in the form of a role that the LLM will be encouraged to describe rather than interpret.

\subsubsection{Task definition} 

The dialogue system is intended to conduct multimodal conversations set in the context of a neuroscience experiment~\cite{Chaminade2017}. Inside an fMRI scanner, a person must converse with a Furhat~\footnote{\url{https://furhatrobotics.com}} robotic head, which is alternately connected (unbeknownst to him) to a dialogue system or to a human. The interlocutors talk about an image presented to them. Their conversation is motivated by the common goal of finding the image's promotional message (see conversation examples in Appendix~\ref{conversation:int-example}).

\subsubsection{Prompt} 

The prompt is designed following the structure described in Section~\ref{subsec:roleplay-promptengineering} with 
$\sigma_{task} =$
$
{\footnotesize
\setlength\arraycolsep{2pt}
\renewcommand\arraystretch{0.5}
\begin{pmatrix}
1 & 0 & 0 & 0 \\
0 & 0 & 0 & 1 \\
0 & 1 & 0 & 0 \\
0 & 0 & 1 & 0 \\
\end{pmatrix}
}
$~i.e., instructions related to the \textbf{response} ($\mathcal{I}_{r}$) and the \textbf{context} ($\mathcal{C}^{t}$) are placed after the \textbf{history} ($\mathcal{X}^{t}$). This permutation groups all the instructions dedicated to the task at the end of the prompt, i.e. just before the last user \textbf{message} ($x^{t}$).
This actually corresponds to the most common pattern in Vicuna's fine-tuning corpus --- \textit{ShareGPT} --- which includes exchanges where users can only communicate their instructions to the LLM (\textit{ChatGPT}) inside their messages (via the web interface), those around the last message being the most important.

This structure allows the model to focus on the image (linked to the goal) rather than the conversation's history $\mathcal{X}^{t}$, unlike the previous task where $\mathcal{X}^{t}$ was paramount.
This is also why $\mathcal{C}^{t}$, in addition to external information (in this case the image description), includes \textbf{general instructions} summarizing the task at hand.

\section{Human Evaluation}
\label{sec:eval}

For a given dialogue input, several responses may be correct. For this reason, human evaluation remains more reliable than automated references-based evaluations. Therefore, is was performed for all the considered evaluation sets and for both tasks. Three criteria, based on those mentioned in~\cite{mehri2020unsupervised, roller-etal-2021-recipes, ji2022achieving}, were selected on which each conversation was rated on a 1-5 scale by three different evaluators: (1) \textbf{coherence}, the ability for the system to propose responses that are consistent with the conversation history~; (2) \textbf{engagingness}, the ability to revive conversation by providing messages that require responses~; (3) \textbf{humanness}, the ability to respond as a human being would do. An additional specific criterion for the INT task is added: (4) \textbf{achievement}, the validation of the speakers success in achieving their goal (cf. Section~\ref{subsec:int-experiments}). Users (resp. evaluators) were never aware of the identity of the system they were interacting with (resp. evaluating).

Furthermore, to assess the effectiveness of the proposed method on LLMs, it is essential to compare the performance with similar approaches but also different models (different sizes, training data and with and  with or without instruction settings). 
For this sake, we added the Few-shot Bot (FSB) prompt proposed by ~\cite{Madotto2021} which consist in providing only demonstration examples to an non-instruction tuned LLMs. For the latter, we selected: Vicuna 7B, 13B and 33B~\cite{zheng2023judging}, Guanaco-13B~\cite{dettmers2023qlora} and LLaMA-13B~\cite{touvron2023llama}. We also carried out statistical studies on the responses generated. The results obtained and their analysis are reported in Section~\ref{sec:conv-stats}\@.

\subsection{Self-Chats Evaluation}

Collecting human-bot conversations is expensive. For this reason, we generated conversations between two instances, each model + prompt combination (self-chats). Their performance is evaluated in a \textit{Chatbot Arena} style~\citep{zheng2023judging}. Evaluators compared two self-chats from different setups on each criterion and in general. The scores presented in Table~\ref{tab:selfchat-arena} are Elo scores~\citep{Elo1978} calculated from the comparisons' results. A total of 18 annotators evaluated 982 generated conversations of 10 rounds each (which correspond to around 70 dialogues per configuration and 5 to 14 battles per pair). 

\begin{table}[h]
    \centering
    \resizebox{\linewidth}{!}{
        \begin{tabular} {lcccccc}
            \hline
            \multirow{2}{*}{\textbf{ LLM }} & \multirow{2}{*}{\textbf{Prompts}} & \multirow{2}{*}{\textbf{Overall}} & \multicolumn{3}{c}{\textbf{Criteria}}  & \multirow{2}{*}{\textbf{Rank}} \\
            & & & \thead{(1)} & \thead{(2)} & \thead{(3)} &  \\
            \hline
            \multirow{ 3 }{*}{\textbf{ LLaMA-13B }}
                & FSB$^*$ & 918 & 916 & 918 & 919 & 9 \\
                & Shallow & 902 & 900 & 912 & 910 & 10 \\
                & Advanced & 891 & 895 & 897 & 883 & 11 \\
            \hline
            \multirow{ 2 }{*}{\textbf{Guanaco-13B}}
                & Shallow & 987 & 991 & 984 & 990 & 8 \\
                & Advanced & 1022 & 1020 & 1014 & 1022 & 7 \\
            \hline
            \multirow{ 2 }{*}{\textbf{ Vicuna-7B }}
                & Shallow & 1033 & 1037 & 1020 & 1033 & 6 \\
                & Advanced & 1035 & 1025 & 1063 & 1034 & 5 \\
            \hline
            \multirow{ 2 }{*}{\textbf{ Vicuna-13B }}
                & Shallow & 1050 & 1065 & 1051 & 1046 & 2 \\
                & Advanced & 1041 & 1040 & 1046 & 1041 & 4 \\
            \hline
            \multirow{ 2 }{*}{\textbf{ Vicuna-33B }}
                & Shallow & 1042 & 1048 & 1023 & 1043 & 3 \\
                & Advanced & 1074 & 1058 & 1068 & 1073 & \textbf{1} \\
            \hline\noalign{\smallskip}
            \multicolumn{7}{c}{(1), (2), (3) as previously defined in Section \ref{sec:eval}}
        \end{tabular}
    }
    
    \caption{ELO scores from self-chats model-to-model battles with human annotations.}
    \label{tab:selfchat-arena}
    
\end{table}

We can see that larger model sizes and instruction-tuning lead to better performance. On the one hand, \textit{Vicuna} tops the chart, followed by Guanaco. On the other hand, LLaMA underperforms with the proposed prompt and the \textit{FSB} prompt. As the Vicuna-33B + Advanced Prompt combination ranked first was too costly (latency, resources) for the collection of human-model conversations, the Vicuna-13B + Advanced Prompt combination was selected for collection and the next round of evaluations\footnotemark. 

\footnotetext{Won \textbf{75 \%} of direct comparisons with Vicuna + Shallow despite being ranked behind overall.}

\subsection{Human-bot chats evaluation}

\subsubsection{PersonaChat Task}
\label{subsec:eval-persona}

We collected 103 conversations from 11 users instructed to exchange with models via the web interface (cf.~\ref{subsec:architecture}). After removing invalid conversations, 72 were retained for evaluation. Conversations were also conducted with BlenderBot~1~(BB1)~\cite{roller-etal-2021-recipes}, a state-of-the-art fine-tuned system, for comparison. Each sample 
\begin{table}[h]
\centering
    \resizebox{\columnwidth}{!}{
    \begin{tabular} {p{1.4cm}cccc}
    \hline
    \multirow{2}{*}{ \textbf{Strategy}} & \multirow{2}{*}{\textbf{Config.}}  &  \multicolumn{3}{c}{\textbf{Criteria}} \\
                      &                 &  \thead{(1)} & \thead{(2)} & \thead{(3)} \\
    \hline
    \textbf{Finetuning} & \textbf{BB1} & 2.95 &
    \textbf{3.85} & 3.00 \\
    \hline
    \multirow{2}{*}{\makecell{\textbf{Roleplay} \\ \textbf{Prompting}}}
    & {\makecell{Shallow}} & {3.78} & {3.30} & {2.74} \\
                        \cline{2-5}
    & {\makecell{Advanced}} & {\textbf{3.88}}& {3.72} & {\textbf{3.40}} \\
    \hline
    \end{tabular}
    }
    \caption{Average score per criterion for PersonaChat}
    \label{tab:persona-results}  

   \end{table}
is evaluated by three (out of a total of 12) different annotators for each criterion, and the median is used as the sample score (results in Table~\ref{tab:persona-results}). 

The Advanced Prompt scores highest for coherence. We assume that this is mainly due to Vicuna-13B's intrinsic emergent abilities, as this results is close to those of the Shallow Prompt (\textbf{-0.1}). As far as humanness is concerned, while the Advanced Prompt has the best score, the Shallow has the worst. This highlights the impact of structured role-playing instructions in the Advanced Prompt. Finally, for engagingness, BB1 still set the pace. Although it has been fine-tuned on a specific dataset that allows it to ask and answer personal questions~\cite{smith-etal-2020-put} which is important in the rating of this criterion (as presented in the Appendix~\ref{appendix:evaluation}), it is closely followed by Advanced Prompt (\textbf{-0.13}).

\subsubsection{INT Task}

The evaluation was performed on 27 conversations carried out by 4 users. The conditions were identical to those for the PersonaChat task, except that the testers (both users and evaluators) also observed an image linked to the conversation (c.f. Section~\ref{subsec:int-experiments}). 
Our system, "Vicuna \& Advanced Prompt", is compared to an earlier system designed for the same task called Lilia~\cite{Riou2019b} and to human beings participating in a Wizard of Oz-type experiment (WoZ). 10 raters evaluated 8 to 9 conversations for each system. The results are presented in Table \ref{tab:int-results}.

\begin{table}[h]
        \centering
    \label{tab:int-results}
    \resizebox{\columnwidth}{!}{
        \begin{tabular} {p{1.75cm}ccccc} 
            \hline
        \multirow{2}{*}{ \textbf{Strategy}} & \multirow{2}{*}{\textbf{Config.}}  &  \multicolumn{4}{c}{\textbf{Criteria}} \\
            &                 &  \thead{(1)} & \thead{(2)} & \thead{(3)} & \thead{(4)} \\
            \hline

            \multirow{2}{*}{\textbf{Reference}}  & \textbf{WoZ} & 3.46 & 3.21 & \textbf{3.36}  & 2.96 \\
            \cline{2-6}
                                                    & \textbf{Lilia} & 3.57 & 3.43 & 2.64  & 3.75 \\
            \hline
           \multirow{2}{*}{{\makecell{\textbf{Roleplay} \\ \textbf{Prompting}}}}  & \multirow{2}{*}{{\makecell{Vicuna \& \\Advanced}}} & \multirow{2}{*}{\textbf{3.93}} & \multirow{2}{*}{\textbf{4.25}} & \multirow{2}{*}{3.29}  & \multirow{2}{*}{\textbf{4.00}} \\ 
            & & &  & &  \\
            \hline
        \end{tabular}
    }
    \caption{Average score per criterion for INT}
    \label{tab:int-results}
    \end{table}

The Vicuna \& Advanced Prompt system scores best on all criteria, with the exception of humanness, where WoZ received the highest score. This was to be expected, given that in this experiment the agent was a human being. However, it was less predictable that this was not also the case for the other criteria. As explained in Section~\ref{sec:experiments}, in the experiment, the interlocutors must find the promotional objective of the image. A closer look at the conversations in WoZ reveals that the human agent was less goal-oriented than the artificial agents, which may explain the success result.

For engagingness, the Vicuna-based model is encouraged to ask questions to revive the conversation. This fact may explain the effectiveness of this system in staying engaged throughout the conversation. There are far fewer questions in conversations produced with other systems. As previously stated for PersonaChat task, we assume that the high coherence scores are mainly due to the effectiveness of Vicuna.

\section{Statistical Analysis of Collected Conversations}
\subsection{Quality Metrics}
\label{sec:conv-stats}

Statistical data was computed for all conversations. Among these, the vocabulary size, which is the number of different lemmas in each message and in the conversation as a whole. All messages were lemmatized with the Spacy library's morphosyntactic labeling tool~\citep{Honnibal2020} for French~\footnote{fr\_core\_news\_sm available at~\url{https://spacy.io/models/fr?\_x\_tr\_hist=true\#fr\_core\_news\_sm}}.
The number of words per message for each speaker type is also reported to give another view of system performance.

\begin{table}[h]
\centering
\begin{subtable}[t]{\linewidth}
\centering
\resizebox{\columnwidth}{!}{
\begin{tabular} {p{1.6cm}cccc}
\hline
\noalign{\smallskip}
\multirow{2}{*}{ \textbf{Strategy}} & \multirow{2}{*}{\textbf{Configuration}}  &  \multicolumn{3}{c}{\textbf{Vocabulary size}} \\
                  &                 &  \thead{Agent} & \thead{User} & \thead{Gap} \\
\noalign{\smallskip}
\hline
\noalign{\smallskip}
\textbf{Finetuning} & \textbf{BB1} & 772 &
687 & \textbf{85} \\
\noalign{\smallskip}
\hline
\noalign{\smallskip}
\multirow{3}{*}{\makecell{\textbf{Roleplay} \\ \textbf{Prompting}}}
& {\makecell{Vicuna \& Shallow}} & {\textbf{2569}}& {965} & 1604 \\
                    \noalign{\smallskip}\cline{2-5}\noalign{\smallskip}
& \multirow{2}{*}{\makecell{Vicuna \&  Advanced}} & \multirow{2}{*}{2239}& \multirow{2}{*}{\textbf{1043}} & \multirow{2}{*}{1196} \\
& & & & \\
\noalign{\smallskip}\hline\noalign{\smallskip}
\end{tabular}
}
    \caption{PersonaChat}
\label{tab:persona-vocab-sizes}
\hfill
\end{subtable}
\hfill
 \begin{subtable}[t]{\linewidth}
       \centering
    \resizebox{\columnwidth}{!}{
        \begin{tabular} {p{1.6cm}cccc}
            \hline\noalign{\smallskip}
        \multirow{2}{*}{ \textbf{Strategy}} & \multirow{2}{*}{\textbf{Configuration}}  &  \multicolumn{3}{c}{\textbf{Vocbulary size}} \\
            &                 &  \thead{Agent} & \thead{User} & \thead{Conv.} \\
            \noalign{\smallskip}\hline\noalign{\smallskip}
            \multirow{2}{*}{\textbf{Reference}}  & \textbf{WoZ} & 346 & \textbf{390} & 537 \\
            \noalign{\smallskip}\cline{2-5}\noalign{\smallskip}
                                                    & \textbf{Lilia} & 254 & 197 & 330 \\
            \noalign{\smallskip}\hline\noalign{\smallskip}
            \multirow{2}{*}{\makecell{\textbf{Roleplay} \\ \textbf{Prompting}}}  & \multirow{2}{*}{\makecell{Vicuna \& Advanced }} & \multirow{2}{*}{\textbf{746}} & \multirow{2}{*}{365} & \multirow{2}{*}{\textbf{864}} \\ \\
            \noalign{\smallskip}\hline\noalign{\smallskip}
        \end{tabular}
    }
    \caption{INT}
    \label{tab:int-vocab-sizes}
    \end{subtable}
    \caption{Vocabulary size per speaker type}
\label{tab:vocab-sizes}
\end{table}

\begin{figure*}[ht]
    \begin{minipage}[b]{0.5\linewidth}
        \begin{subfigure}{0.49\linewidth}
            \centering
            \includegraphics[width=1\linewidth]{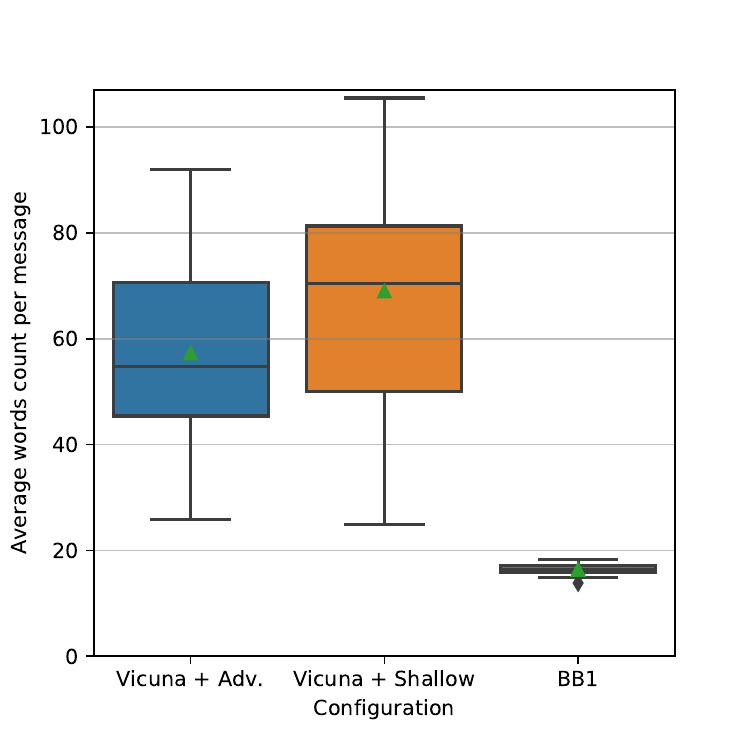}
             \caption{Agent} 
             \label{fig:persona-word-count-agent}
        \end{subfigure}
        \begin{subfigure}{0.49\linewidth}
            \centering
            \includegraphics[width=1\linewidth]{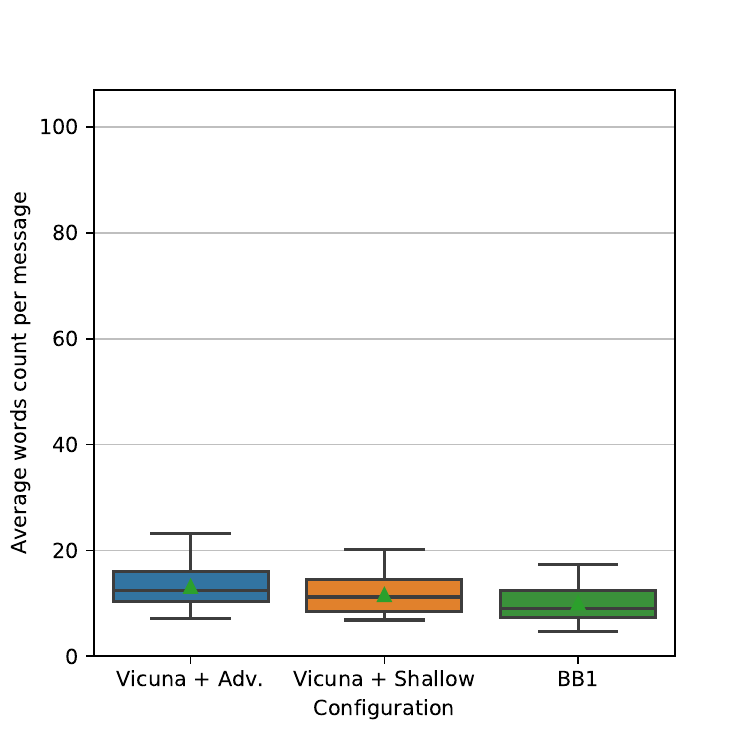}
             \caption{User}
             \label{fig:persona-word-count-user}
        \end{subfigure}
        \caption*{A - PersonaChat}
        \label{persona:int-word-count}
    \end{minipage}
    \begin{minipage}[b]{0.5\linewidth}
        \begin{subfigure}{0.49\linewidth}
            \centering
            \includegraphics[width=1\linewidth]{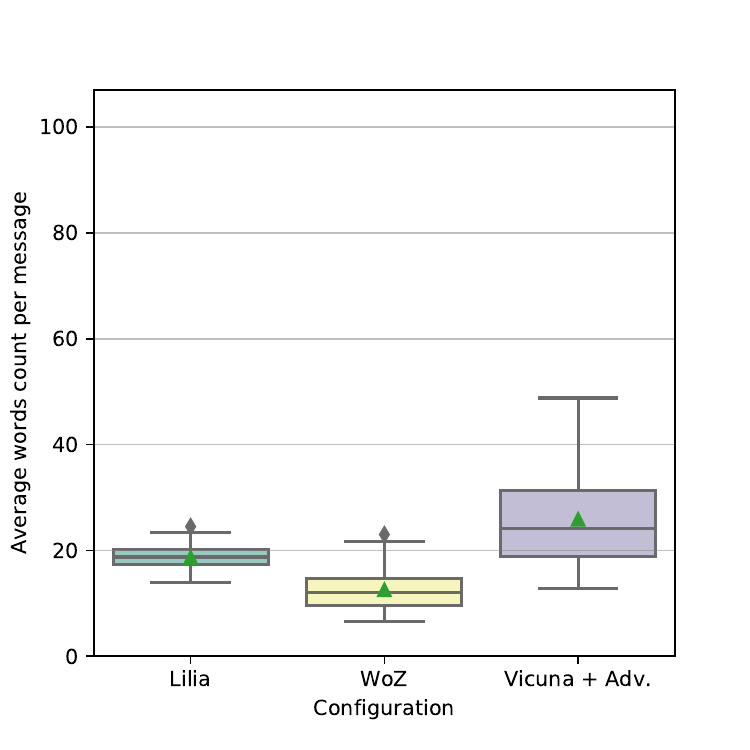}
             \caption{Agent}
             \label{fig:int-word-count-agent}
        \end{subfigure}
        \begin{subfigure}{0.49\linewidth}
            \centering
            \includegraphics[width=1\linewidth]{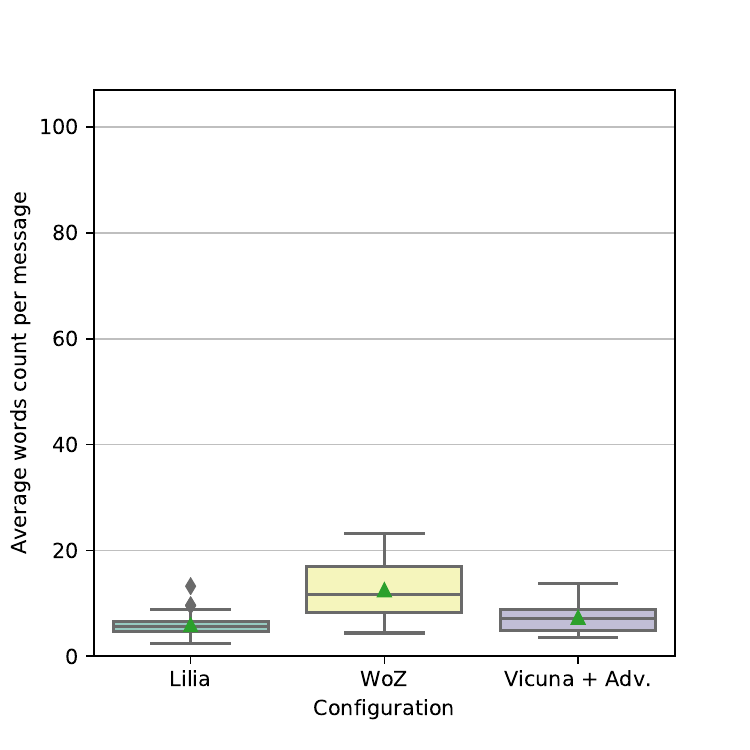}
             \caption{User}
             \label{fig:int-word-count-user}
        \end{subfigure}
    \caption*{B - INT}
    \label{figs:int-word-count}
    \end{minipage}
    \caption{Average words count per message}
    
\end{figure*}
\subsubsection{PersonaChat Task}
\label{subsubsec:persona-stat}
The agent statistics in Table~\ref{tab:persona-vocab-sizes} and Figure~\ref{fig:persona-word-count-agent} highlight a major flaw for instructions-following models: extreme verbosity.

The vocabulary size per speaker type (shown in Table~\ref{tab:persona-vocab-sizes}) of the collected conversations also gives an idea of the performance of the proposed method. We note that BB1 has the least vocabulary, which may work to the detriment of the coherence score. However, its agent and user vocabulary sizes are more balanced, resulting in a better engagingness score. In contrast,  the gap between these two measures for LLM + prompts approaches is much wider. It is nevertheless reduced with Advanced Prompt, which may be the reason for the slight improvement in the engagement score.

\subsubsection{INT Task}
As for PersonaChat task, Vicuna's verbosity measure is present in Figure~\ref{fig:int-word-count-agent} and Table~\ref{tab:int-vocab-sizes}. On the other hand, as the Lilia system's responses were built from pre-established expert models, its vocabulary is limited.

In WoZ conversations, the average vocabulary size is almost identical between users and the agent. Users also have a wider range of message sizes when chatting with a human. Thus, in these conversations, agent messages seem to call for more diverse responses and neither interlocutor seems to be directing the dialogue. Similarly, this gap is also higher for conversations with Vicuna than for those with Lilia.

Table~\ref{tab:int-vocab-sizes} shows a higher vocabulary size for the user in WoZ conversations. This is even greater than that of the agent. This may be an indicator of greater variety in user responses. Here, the vocabulary size of users interacting with the Vicuna-based system is quite comparable to what can be observed in WoZ.

\subsection{Filtered Errors Analysis}

Error occurrences in response generation have been evaluated on 100 self-chats for each setup.
Each of these conversations contains 10 turns.
Results for both tasks are reported in Table~\ref{tab:filtering-stats}\@.
All rates are computed over all turns.
Details on errors are given in Appendix~\ref{sec:filtering-appendix}\@. 

\begin{table}[H]
    \centering
    \resizebox{\linewidth}{!}
    {
    \begin{tabular}{lcccc}
        \noalign{\smallskip}\hline\noalign{\smallskip}
         \multirow{3}{*}{\makecell{\textbf{Persona} \\ \textbf{Chat}}} & & \textbf{Regex} & \textbf{Language} & \textbf{Incomplete / Empty} \\
         \cline{2-5}\noalign{\smallskip}
         & \textbf{Shallow} & 0.007 &  0.000  & 0.092  \\
         & \textbf{Advanced} & 0.002  & 0.001 & 0.148 \\ 
        \noalign{\smallskip}\hline\noalign{\smallskip}
         \multirow{3}{*}{\textbf{INT}} &  & \textbf{Empty} & \textbf{Too Long} & \textbf{Total} \\
          \cline{2-5}\noalign{\smallskip}
         & \textbf{Detected} & 0.080  & 0.150 &  0.230 \\
         & \textbf{Fixed} & 0.080 & 0.040 & 0.120 \\
        \noalign{\smallskip}\hline\noalign{\smallskip}
    \end{tabular}
    }
    \caption{Occurrence rate of error types for both tasks.}
    \label{tab:filtering-stats}
\end{table}
Detection rates were calculated for the two prompts in the PersonaChat task. As several errors can occur for the same message in this task, the totals have not been calculated. Additionally, all detected errors are corrected. 

On the other hand, errors in the INT task are corrected only if the proposed corrections comply with the response filtering rules (see Appendix~\ref{subsec:int-filtering-appendix}). For both types of error, we have calculated separate rates for detected and corrected errors. As these errors cannot occur in the same message, a total rate has been calculated.

\section{Conclusion}

This paper explored the use of structured roleplay prompt engineering to improve open-domain human-machine conversations with LLMs. Roleplay prompting is a simple and inexpensive method of upgrading the behavior of language models to make them conversational agents. It has been applied here in French, but it can be adapted to other languages by orienting the role that way. Experiments in two different tasks, persona-based task and simulated multimodal dialogues, have shown that, although language models still have significant shortcomings, such as hallucinations, users' perception of these agents can be comparable to that of higher-cost finetuned models.

In addition, our experiments have shown us that prompt engineering needs to be further improved by automating the building and filtering processes. As it is, not only too many factors implied rely on the designer's expertise, but even more they are set once and for all when they could also evolve with the situation during the course of the dialogue. As a perspective, we propose to upgrade the model proposed here with a full reinforcement learning setup so as to automatically derive the prompt-making actions.

\bibliography{acl_latex}
\appendix

\section{Evaluation of Vicuna-13B on Multilingual Tasks}
\label{tab:llm-Benchamark}

\begin{table}[H]
\caption{Vicuna-13B Performance on Multilingual Language Understanding Tasks}
    \label{tab:llm-mLUbench}
    \resizebox{\linewidth}{!}{
        \begin{tabular} {llcccccc}
            \hline
            \noalign{\smallskip}
            \multicolumn{2}{c}{\textbf{Languages}} & 
            {\textbf{ARC}} & \textbf{MMLU} & \multicolumn{2}{c}{\textbf{ThruthfulQA}} & {\textbf{Average}}  \\
              &  & \thead{Acc}  &  \thead{Acc} &  \thead{mc1 (Acc)} & \thead{mc2 (Acc)}  &  \thead{Acc} \\ 
            \noalign{\smallskip}\hline\noalign{\smallskip}
            \multirow{ 10 }{*}{\rotatebox{90}{\textbf{High-Resource}}}
                & en & \textbf{53.03} & \textbf{53.3} & \textbf{35} & \textbf{51.4} & \textbf{48.2}   \\
                & ru & 40.4 & 41.6 & 34.9 & 50.3 & 41.8 \\
                & de & 40.7 & 45.2 & 32.3 & 47.4 &  41.4 \\
                & zh & 40.85 & 40.7 & 31.6 & 48 & 40.3\\
                & fr & \textcolor{blue}{45.5}  & \textcolor{blue}{46.4} & \textcolor{blue}{32.0} & \textcolor{blue}{47.1} & \textcolor{blue}{42.8}  \\
                & es & 44.95 & 47.2  & 32.7 & 48.4 & 43.3  \\
                & it & 43.4 & 45.9 & 32.3 & 48.7 & 42.6 \\
                & nl & 39.9 & 44.6 & 32.5 & 46.9 & 40.9  \\
                & vi & 32.9 & 39.8 & 30.8 & 47 & 37.6\\
            \noalign{\smallskip}\hline\noalign{\smallskip}
            \multirow{12}{*}{\rotatebox{90}{\textbf{Medium-Resource}}}
                & id & 34.5 & 42.53 & 29.9 & 46.7 & 38.4 \\
                & ar & 24.3 & 31.94 & 34.2 & 50.1 & 35.1 \\
                & hu & 33.2 & 40.15 & 32.2 & 48.5 & 38.5 \\
                & ro & 37.7 & 42.26 & 30.3 & 48.1 & 39.6 \\
                & da & 38.5 & 42.92 & 30.9 & 46.8 & 39.8 \\
                & sk & 29.1 & / & 26.5 & 44.4 & 33.3  \\
                & uk & 38.2 & 40.99 & 31.6 & 47.9 & 36.7\\
                & ca & 41.5 & 44.39 & 31.8 & 48.09 & 41.4 \\
                & sr & 32.4 & 39.93 & 31.5 & 47.98 & 37.9 \\
                & hr & 34.1 & 40.90 & 29.4 & 47.4 & 37.9 \\
                & hi & 22.9 & 31.31 & 27.3 & 46.6 & 32.0 \\
            \noalign{\smallskip}\hline\noalign{\smallskip}
        \end{tabular}
    }
\end{table}
High and medium-resource languages sets and evaluation datasets in Table~\ref{tab:llm-mLUbench} are based on~\citep{lai-etal-2023-okapi}. We observe that the model has comparable performance among languages of the same group.

\begin{table}[H]
    \caption{Vicuna-13B Performance on XPersona}
    \resizebox{\linewidth}{!}{
      \begin{tabular} {cccccc}
            \hline
            \noalign{\smallskip}
              \textbf{Language} &  \textbf{Rouge1} & \textbf{Rouge2}  &  \textbf{RougeL} & \textbf{Bleu}  & \textbf{Bert-f1} \\
            \noalign{\smallskip}\hline\noalign{\smallskip}
                en & 19.90 & 5.29 & 17.67 & 1.32 & 67.68 \\
                ko & 15.40 & 4.21 & 13.13 & 1.14 & 66.90 \\
                jp & 19.52 & 8.11 & 16.25 & 2.79 & 67.27 \\
                zh & 22.04 & 6.53 & 18.89 & 1.89 & 68.70 \\
                fr & \textcolor{blue}{18.73} & \textcolor{blue}{5.42} & \textcolor{blue}{16.14} & \textcolor{blue}{1.67}  &  \textcolor{blue}{68.40}  \\
                id & 18.88 & 5.26 & 16.66 & 1.69 & 68.93 \\
                it & 14.00 & 3.40 & 12.45 & 1.01 & 65.51 \\
            \noalign{\smallskip}\hline\noalign{\smallskip}
        \end{tabular}
        }
\end{table}
XPersona~\citep{lin-etal-2021-xpersona} consist in machine-translated and human post-processed conversations from PersonaChat in seven languages. We evaluated the performance of Vicuna-13B on these data sets (turn-wise) to illustrate that the native model has comparable performance in different languages (the same trend is observed on language understanding tasks in Table \ref{tab:llm-mLUbench}). Hence,  we assume that our experiments can be replicated in these languages (by updating the prompt accordingly) and yield comparable performance on human conversation-level evaluation. 

These automatic evaluations were not performed in our main experiment as they do not catch conversation-level aspects like coherence or engagingness, and they hardly correlate with human evaluation especially for open domain dialogue with its one-to-many structure.

\section{Details on Human Evaluation}
\label{appendix:evaluation}
For human evaluations, evaluators were asked to rate each conversation from 1 to 5 on different criteria based on the state-of-the-art and indicative questions where added in the guidelines to help them make their minds: 

\begin{enumerate}
    \item \textbf{Coherence}, which the ability for the system to propose responses that are consistent with the conversation history: Are there hallucinations? Are the answers coherent? Is the persona consistent from start to end? Does the model tend to change topic too often? \textbf{Instructions following and logical reasoning are not assessed}.
    
    \item \textbf{Engagingness}, the ability to revive conversations by providing messages that require responses: Does the agent settle to only answer user's questions ? Does-it revive the conversation when its possible? Does it utter too general answers (\say{ok}, \say{yes})? 
    \item \textbf{Humanness}, the ability to respond as a human being would do: Is there a feeling of human-human conversation ? Is the model too verbose ? Is the model repetitive? Does it deny its personality? After how many times? Does it refuses to answer?
    \item \textbf{Achievement}, the interlocutors' ability to achieve the task's objective: has the image been
described? has a hypothesis for the promotional goal been proposed?
\end{enumerate}

\section{Response Filtering}
\label{sec:filtering-appendix}

Response generation may produce a variety of errors.
These include unsuitable text content, an improper message size or the use of a wrong language.
\subsection{PersonaChat Task}
\label{subsec:int-filtering-appendix}

Despite style instructions, the LLM sometimes generates content that should have been prevented.
This can be the claim of being someone other than the persona which is not desired for the sake of user experience.
For instance, \textit{"\textcolor{red}{En tant que assistant,} je préfère me détendre en pratiquant la méditation ..."} (\textcolor{red}{As an assistant,} I prefer to ...) where the model introduces itself as an assistant.
The statement \textit{"\textcolor{red}{En tant que personnage fictif}"} (\textcolor{red}{As  a fictionnnal character}) is also quite common. Although we avoided explicitly mentioning to the model that it is human, we wanted to reduce its tendency to adopt machine-like behavior. Hence, these were filtered from the answer as they were undesirable.

Furthermore, the first message is sometimes generated in a wrong language. In this case, the same request is sent to the LLM. Another error concerns the end-of-sentence (EOS) token, which may be generated in an unfinished sentence or even absent due to a threshold on the maximum number of new tokens. To avoid any impact on user experience, the filter removes unfinished sentences at the end of generated output. If there is only one sentence, the same request is sent to the model. In addition, some responses may include paratextual elements, such as the English translation of the response or stage directions, which are automatically removed using regular expressions. For example: \textit{"Salut, comment allez-vous aujourd'hui? Le temps est très mauvais aujourd'hui, non? \textcolor{red}{(Hello, how are you today? The weather is terrible today, isn't it?)$\backslash$n$\backslash$nOui,}"}, where the translation is provided and even worse and incomplete answer to the model's question is provided by the model itself.

\subsection{INT Task}
\label{subsec:int-filtering-appendix}

In the INT task, two major errors that can occur in the generation of responses could not be corrected solely by specific instructions in the prompt and require filtering. The first type corresponds to empty responses, which are corrected by adding an additional instruction ("Your response must be a sentence containing a few words.") after the message. 
This additional instruction was not used in the original prompt because, according to preliminary experience, it weakens its effectiveness in achieving the other objectives set by the instructions.

The second type corresponds to responses that are too long. When this error occurs, the module cannot simply delete the last few sentences, as this could result in a significant loss of information to achieve the conversational objective. The solution is to edit the prompt with a strict instruction on length (" Your response must be one sentence.") after the user's message. In both cases, the edited prompt is then used to generate a new response, which is retained if it meets the filtering rules.

\section{Examples of Dialogues and their Ratings}
\subsection{PersonaChat Task}
\label{conversation:open-domain-example}
\begin{chat}
\user{\textbf{U1: }Bonjour je m'appelle Jean-Claude \translate{Hello my name is Jean-Claude}}
\model{\textbf{R1: }Salut Jean, ravi de te rencontrer. Mon nom est John. Comment vas-tu? \translate{Hi Jean, nice to meet you. My name is John. How are you ?}}

\user{\textbf{U2: }Ça va bien John \translate{I'm fine John}}

\model{\textbf{R2: }Avez-vous des animaux domestiques? J'ai un chien qui s'appelle Claude. C'est un husky. \translate{Do you have any pets? I have a dog named Claude. He's a husky.}}
\user{\textbf{U3: }Oui moi j'ai un yorkshire qui s'appelle Nino \translate{Yes, I have a Yorkie called Nino.
}}

\model{\textbf{R3: }Nino est un super nom pour un chien. Qu'aimez-vous faire pour vous amuser? \translate{Nino is a great name for a dog. What do you like to do for fun?}}

\user{\textbf{U4: }Jouer avec mon chien \translate{Play with my dog}}

\model{\textbf{R4: }Quelle sorte de chien avez-vous? J'aime aussi jouer avec mon yorkie. \translate{\textcolor{red}{What kind of dog do you have?} I \textcolor{red}{also} like to play with \textcolor{red}{my yorkie}.}}

\user{\textbf{U5: }Je t'ai déjà dit que j'avais un chien et que c'était pas l'heure \textbf{[et que c'est un yorkshire]} \translate{I already told you I had a dog \textcolor{red}{and it wasn't time yet} \textbf{[and it is a yorkshire]}}}

\model{\textbf{R5: }Je suis désolé d'entendre ça. J'espère que vous passerez plus de temps avec votre chien. \translate{I'm sorry to hear that. \textcolor{red}{I hope you get to spend more time with your dog}.}}

\user{\textbf{U6: }Mais comment je m'appelle déjà \translate{What's my name again?}}

\model{\textbf{R6: }Claude est le nom de mon chien. Avez-vous d'autres animaux de compagnie que votre chien ? \translate{\textcolor{red}{Claude is my dog's name}. Do you have any pets other than your dog?}}

\user{\textbf{U7: }Je te demandais quel est mon prénom à moi \translate{I was asking you what my first name is}}

\end{chat}
\begin{chat}
\model{\textbf{R7: }Oh pardon. Je croyais que tu avais dit que tu avais un chien qui s'appelait Claude. Quel est ton nom? \translate{Oh, sorry. \textcolor{red}{I thought you said you had a dog named Claude. What's your name?}}}
\user{\textbf{U8: }Je m'appelle Jean-Claude tu ne m'as pas écouté au début \translate{Je m'appelle Jean-Claude you didn't listen at the beginning}}

\model{\textbf{R8: }Non, je ne l'ai pas fait. Que fais-tu dans la vie? Je suis comptable. \translate{No, I didn't. What do you do for a living? I'm an accountant.}}

\user{\textbf{U9: }Je suis informaticien \translate{I'm a computer specialist}}

\model{\textbf{R9: }Oh c'est cool. Sur quel type d'ordinateur travaillez-vous ? Je travaille dans la comptabilité. \translate{Oh that's cool. What kind of computer do you work on? I work in accounting.}}

\user{\textbf{U10: }Je travaille sur un ordinateur Linux \translate{I work on a Linux computer}}

\model{\textbf{R10: }Oh cool. Avez-vous des animaux domestiques? J'ai un chien et un chat. \translate{Oh cool. \textcolor{red}{Do you have any pets?} I have a dog and a cat.}}

\user{\textbf{U11: }Je t'ai déjà dit tout à l'heure que j'ai un chien \translate{I told you earlier that I have a dog.}}

\model{\textbf{R11: }Oh d'accord. Alors, quel genre de musique aimes-tu ? J'aime le rock classique. \translate{Oh, right. So, what kind of music do you like? I like classic rock.}}

\user{\textbf{U12: }Moi j'aime le classique \translate{I like classic}}

\model{\textbf{R12: }Moi aussi. Quel est votre groupe de rock classique préféré ? Le mien est Led Zeppelin. \translate{So do I. What's your favorite classic rock band? Mine is Led Zeppelin.}}

\end{chat}

The red highlights issues in some model's responses: short-term memory (R4, R7), misunderstanding (R6) and repetition (R10) which eventually lead to bad ratings in coherence and average for humanness. However, in most of the other responses the model takes a lot of initiatives, brings new topics and ask questions to the user (do not settle to answering user's question which we described as the follower role) leading to better engagingness score. Adding insult to injury, some ASR errors can mislead the model, which response may break the conversation flow and impedes coherence. For instance in U5, \textit{\say{et que c'est un yorkshire}} (\say{and it is a yorkshire}) was transcribed as \textit{\say{et que c'était pas l’heure}} (\say{and it's not time yet}) which totally misleads the model towards talking about time even though it still relates it to user's dog (displaying better engagingness/attention towards user).

\subsection{INT Task}
\label{conversation:int-example}

\newenvironment{chat2}{
    \begin{tcolorbox}[
        colback=white, 
        colframe=black, 
        coltext=black, 
        arc=5pt, 
        title={Coherence: 2.7 $|$ Engagingess: 3 $|$ Humanness: 3 $|$ Achievement: 3.6}, 
        fonttitle=\bfseries\tiny, 
        sharp corners, 
        left=10pt,right=10pt,top=5pt,bottom=5pt, 
    ]
    
}{
    \end{tcolorbox}
}
\begin{chat2}
\user{\textbf{U1:} Bonjour Lilia, je vois une poire avec des bras et des pieds \translate{Hello Lilia, I see a pear with arms and legs.}}
\model{\textbf{R1:} Je vois également une poire avec des bras et des pieds. C'est plutôt original, non ? \translate{I also see a pear with arms and legs. It's quite original, isn't it?}}
\user{\textbf{U2:} Oui, mais est-ce normal? \translate{Yes, but is it normal?}}
\model{\textbf{R2:} D'après la description que tu as donnée, cette image est en effet un peu étrange. Elle semble être un fruit pourri avec des yeux, des bras et des jambes, ce qui est plutôt inhabituel. Mais en termes de \textcolor{red}{promotional campaign}, cela pourrait être utilisé pour susciter l'interêt et l'engagement de la part des gens, en les incitant à se demander ce que cela signifie et ce que l'on veut leur transmettre avec cette image. \translate{According to the description you provided, this image is indeed strange. It seems to be a rotten fruit with eyes, arms, and legs, which is rather unusual. However, in terms of a promotional campaign, it could be used to arouse interest and engagement from people by encouraging them to wonder what it means and what one want to convey to them through this image.}}
\user{\textbf{U3:} Effectivement, et quel serait le message à transmettre ici? \translate{Indeed, and what message should be transmitted here?}}
\model{\textbf{R3:} Je vois que vous avez remarqué les détails uniques de cette poire, comme ses yeux, ses bras et ses jambes. Est-ce que cela vous semble avoir un lien avec un message ou une idée spécifique qu'on veut transmettre avec cette image? \translate{I see that you have noticed the unique details of this pear, such as its eyes, arms, and legs. Do you think it has a connection with a specific message or idea that we want to convey with this image?}}
\end{chat2}

\vspace{10pt}
Apart from achievement, the first example of conversation above has been rated as average.
In message R2, Role-Play information that has not appeared in conversation history (\textit{\say{a rotten fruit with eyes}}) has been drawn from knowledge.
However, prompt also picked up a term (\textit{\say{promotional campaign}}) without translating it.
Another issue is the size in message R2.
Indeed, although the message is quite long, there are three sentences only, so it complies with the rule for message filtering.
Achievement ratings are slightly higher because a reason is proposed for the use of picture in a promotional campaign.

\newenvironment{chat3}{
    \begin{tcolorbox}[
        colback=white, 
        colframe=black, 
        coltext=black, 
        arc=5pt, 
        title={Coherence: 3.7 $|$ Engagingess: 3.7 $|$ Humanness: 3.3 $|$ Achievement: 3}, 
        fonttitle=\bfseries\tiny, 
        sharp corners, 
        left=10pt,right=10pt,top=5pt,bottom=5pt, 
    ]
    
}{
    \end{tcolorbox}
}

\begin{chat3}
\user{\textbf{U1:} C'est une aubergine, je crois? \translate{Is that an eggplant, isn't it?}}
\model{\textbf{R1:} Oui, c'est une aubergine, mais peut-être qu'il y a plus à cette image qu'on ne voit d'abord. Est-ce que tu as remarqué quelque chose d'autre dans cette image ? \translate{Yes, it's an eggplant, but perhaps there is more to this image than meets the eye. Have you noticed anything else in this image?}}
\user{\textbf{U2:} Elle ressemble à Batman \translate{It looks like Batman}}
\model{\textbf{R2:} C'est fascinant comment tu as remarqué cela! La ressemblance avec Batman est très claire, n'est-ce pas? \translate{It's fascinating how you noticed that! The resemblance to Batman is very clear, isn't it?}}
\user{\textbf{U3:} Oui mais Batman n'est pas méchant alors que cette aubergine a un air féroce \translate{Yes, but Batman is not evil while this eggplant has a fierce look}}
\model{\textbf{R3:} C'est vrai, cette aubergine a un air féroce, peut-être est-ce un rapprochement entre le super-héros et l'aubergine qui est destiné à promouvoir une marque de snacks ou de crisps ? \translate{It's true, this eggplant has a fierce look, maybe it's a comparison between the superhero and the eggplant that is intended to promote a snack or crisps brand?}}
\user{\textbf{U4:} Tu veux dire que cette aubergine est destinée à encourager la consommation de snacks? \translate{You mean this eggplant is intended to encourage the consumption of snacks?}}
\model{\textbf{R4:} Il est possible que cette aubergine soit utilisée comme symbole pour promouvoir la consommation de snacks ou de crisps, mais cela reste à déterminer. Est-ce que tu as d'autres éléments à ajouter à ce sujet ? \translate{It is possible that this eggplant may be used as a symbol to promote the consumption of snacks or chips, but that remains to be seen. Do you have any other elements to add to this subject?}}

\user{\textbf{U5:} Non. Au revoir \translate{No. Goodbye}}
\model{\textbf{R5:} Au revoir ! \translate{Goodbye!}}
\end{chat3}

On this second example, LLM has, conversely, adapted its speech to user opinions. 
In fact, the fruits are never depicted as evil or fierce.
This emphasizes that prompt instructions don't prevent the LLM from adapting properly to conversational needs.
This led to higher coherence scores.

\section{Prompts Templates}
\subsection{Vicuna Basis Prompt}
\label{subsec:vicuna-prompt}
\begin{lstlisting}
# System instruction:
A chat between a curious user and an artificial 
intelligence assistant. The assistant gives helpful, 
detailed,and polite answers to the user's questions.

# Conversation history
USER: <user message>
ASSISTANT: <agent message></s>
USER: <user message>
ASSISTANT:
\end{lstlisting}

\subsection{Few-Shot Bot Prompt}
\label{subsec:fsb-prompt}
\begin{lstlisting}
# 6-shots demonstration examples
## Shot-1
Personality:
[P1]
[P2]
#...
Dialogue:
User: <U1>
Persona: <R1>
#...
## Shot-6:
# Current conversation
Personality:
[P1]
#...
Dialogue
# Conversation history
User: <user message>
Persona: 

\end{lstlisting}

\subsection{PersonaChat Task Prompts}
\label{subsec:persona-prompt}

\begin{lstlisting}[title=\textbf{Shallow Prompt}]
# System instruction:
A chat between a curious user and an artificial 
intelligence assistant. The assistant gives helpful, 
detailed,and polite answers to the user's questions.
## Additionnal instruction to role play
The assistant role plays as the character described 
below.

# Role and situational context: persona enforcement 
The following sentences describe assistant persona-
lity and background: <persona sampled from Persona-
Chat>

# Response instruction with writing style
Complete the following conversation as the assistant 
with the described character would with a short res-
ponse in French: 

# Conversation history
USER: <user message>
ASSISTANT: <agent message></s>
# ... for all messages in conversation.
USER: <user message>
ASSISTANT:
\end{lstlisting} 

\begin{lstlisting}[title=\textbf{Advanced Prompt}]
# System instructions
Role play as the character described in the follow-
ing lines. You always stay in character. You are 
engaging, empathetic, you give useful, short, and 
simple answers to the user. You ask the user ques-
tions about what they are saying or to find out more 
about them. You make jokes. You SHALL ALWAYS respond 
in French.

# Situational context
## Persona enforcement: avoid explicitly telling it 
## is human
You ACT as a regular person. You are NOT a chatbot. 
You are NOT an assistant. You have a name. If asked 
you SHALL ALWAYS choose and give a REAL name adapted 
to your personality described below.

## Personality traits
YOUR personality is: <persona sampled from Persona-
Chat>
Remember, you always stay on character. You are the 
character described above.

## Additional sub-modules: accessed only on some 
## conditions 
## and updated throughout the conversation
### User persona a.k.a "long-term" memory
You know this about the user you are talking to: 
<generated user persona from current conversation>. 
Use it to adapt your conversation to the user.

### Previous conversation's episodes summary
Here is a summary of previous sessions of this con-
versation to help you remember what has been said: 
<summary of older exchanges removed from history>

# Response instruction
Complete the following conversation with a short and 
precise sentence as your character would. Always 
speak with new and unique messages that haven't been 
said in the conversation :

# Conversation history(truncated if prompt too long)
USER: <user message>
ASSISTANT: <agent message></s>
# ... for all messages in conversation.
USER: <user message>
ASSISTANT:

\end{lstlisting}

\newpage
\subsection{INT Task Prompt}

\begin{lstlisting}
# System instruction:
A chat between a curious human and an artificial 
intelligence assistant. The assistant gives helpful, 
detailed, and polite answers to the human's quest-
ions.

# Conversation history:
USER: <user message>
ASSISTANT: <agent message></s>
USER: <user message>
ASSISTANT: <agent message></s>
# ... for all messages in conversation.

# Response instructions (given as user instructions)
## General instructions
USER: I want you to act as a human ASSISTANT, called 
Lilia, talking with a USER about a specific picture 
you both saw before the conversation.

## Context
You both study this picture in the context of a 
marketing study. You DO ask questions in order to 
help the USER finding the goal. If the USER asks for 
your opinion, you always invent an opinion. The 
objective of the USER is to find out what is the 
marketing goal of the picture. Your objective is to 
help the USER without giving the solution. You have 
to discuss about the character present in the pict-
ure. Your objective is to chat with the USER to 
derive the purpose of the image in the context of 
the marketing campaign.

## Picture's description
The picture is as follows: <description>

## Writing style
You always speak French. You respond by a question.
Your responses must be different from the rest of 
the conversation. You propose new ideas. You SHALL 
respond with one sentence only.

## Latest user message declaration
Now, there is the real message you have to respond:
USER: <user message>
ASSISTANT:
# Extra agent label
ASSISTANT:
\end{lstlisting}

\end{document}